\documentclass[a4paper]{article}

\usepackage{INTERSPEECH_v2}
\usepackage{amssymb}
\usepackage{graphicx}
\usepackage{amsmath,graphicx}
\usepackage{tikz}
\usepackage{blindtext}
\usepackage{graphicx}
\usepackage{tikz}
\usetikzlibrary{arrows.meta}
\usetikzlibrary{shapes,decorations,arrows,shadows,shapes.geometric}
\usepackage[colorlinks]{hyperref}
\usepackage[colorinlistoftodos]{todonotes}
\usepackage{smartdiagram}
\usetikzlibrary{chains}
\usepackage{amsmath,bm,times}
\usepackage{amsmath}
\usetikzlibrary{mindmap,trees}
\usetikzlibrary{trees}
\usepackage{mathtools}

\usepackage{algorithm}
\usepackage{algorithmic}
\usepackage{tikzscale}
\usepackage{filecontents}

\title{Use of Knowledge Graph in Rescoring the N-best List in Automatic Speech Recognition}
\name{Ashwini Jaya Kumar$^1$, Camilo Morales$^2$, Maria-Esther Vidal$^1$, Christoph Schmidt$^1$, S\"oren Auer$^{1,2}$}
\address{
  $^1$Fraunhofer IAIS, Schloss Birlinghoven, Sankt Augustin, Germany\\
  $^2$University of Bonn, Bonn, Germany}
\email{\footnotesize \ ashwini.jaya.kumar@iais.fraunhofer.de}

\begin{document}

\maketitle
\begin{abstract}

With the evolution of neural network based methods, automatic speech recognition (ASR) field has been advanced to a level where building an application with speech interface is a reality. Inspite of these advances, building a real-time speech recogniser faces several problems such as low recognition accuracy, domain constraint and out-of-vocabulary words. The low recognition accuracy problem is addressed by improving the acoustic model, language model, decoder and by rescoring the N-best list at the output of decoder. We are considering the N-best list rescoring approach to improve the recognition accuracy. Most of the methods in literature uses the grammatical, lexical, syntactic and semantic connection between the words in a recognised sentence as a feature to rescore. In this paper, we have tried to see the semantic relatedness between the words in a sentence to rescore the N-best list. Semantic relatedness is computed using TransE~\cite{bordes2013translating}, a method for low dimensional embedding of a triple in a knowledge graph. The novelty of the paper is the application of semantic web to automatic speech recognition.

\end{abstract}


\section{Introduction}

Automatic speech recognition (ASR) is a system to convert spoken signal into it's transcription (text). The performance of ASR is determined by how accurately the spoken words are recognised. The recognition accuracy is vital in any application where speech interface is required. Considering a question answering application, e.g.``What is the capital of Germany?'', in this sentence if an important entity is misrecognised by ASR, the whole meaning of the sentence is changed and thus affects the interconnected systems. An ASR system mainly consists of three main components, acoustic model, language model and decoder. Acoustic model is a sub-system which statistically connects the features extracted from speech signal to it's phoneme representation. Language model assigns probability for sequence of words. It helps to constrain the search among alternative words. While the decoder helps to combine both the acoustic score and language model score to decode the sequence of words (or characters or phones). In the literature, recognition accuracy is improved by improving the acoustic model, the language model, at the decoder and at the output of ASR by rescoring the N-best list. N-best list is a list of hypothesis of a sentence by a recognition system.  Subsequent re-ranking of the N-best list using various knowledge sources is called rescoring. Let $S^1 = {s_1, s_2, s_3,\dots,s_n,\dots,s_N}$ be the $N$ different form of a sentence $S^1$ at the output of a speech recogniser. These best possible sentences are called N-best sentences.

A knowledge graph is a connected graph of triples in $<subject, predicate, object>$ form with entities as subjects/objects and predicate as the property/relation with which the entities are connected. A triple is a logical representation of the words and it's relations.e.g. $<subject, predicate, object>$. A relation is a property with which the two entities are connected e.g. daughter of, employee of, president of, etc,.The words in a triple can be an entity, a real, discrete unit e.g. person, organisation, place, etc,.The use of external knowledge sources like knowledge graph to improve the recognition accuracy is not tested in any of the previous methods. In this paper, we are explaining the use of external knowledge resource to rescore the N-best list. We are using dbpedia knowledge graph as the external knowledge source. 

% Define the layers to draw the diagram
\pgfdeclarelayer{background}
\pgfdeclarelayer{foreground}
\pgfsetlayers{background,main,foreground}

% Define block styles used later

\tikzstyle{sensor}=[draw, fill=blue!20, text width=5em, 
    text centered, scale=0.8, minimum height=2em]
\tikzstyle{ann} = [above, text width=5em, text centered]
\tikzstyle{wa} = [sensor, text width=8em, fill=red!20, 
    minimum height=6em, rounded corners]
\tikzstyle{sc} = [sensor, text width=10em, fill=red!20, 
    minimum height=7em, rounded corners]

Entity linking is a method to link the triple (entities/relations) in a sentence to the RDF molecules of a knowledge graph. The N-best list of sentences and the knowledge graph is connected by linking the entities/relation in a sentence to the triple units in the knowledge base. Dbpedia spotlight, a tool to annotate the mentions of DBpedia is used here for entity/relations linking in the N-best sentences to the Dbpedia knowledge graph. The output of Dbpedia spotlight contains the link (uri) for a given entity/relations. e.g. for entity``Barrack\_Obama", Dbpedia uri link looks like, http://dbpedia.org/page/Barack\_Obama. Dbpedia spotlight only gives the linking information. Next, we are trying to see whether these entities and relations are connected to each other in knowledge graph. If they are connected, then the difference of TrasE embeddings between that RDF molecule of $entity_1$ and RDF molecule of $entity_2$ is zero. In other words, we are applying modified viterbi algorithm across the RDF molecues of entities/relations in a sentence. Difference between TransE embeddings is used as a cost function. To get the TransE embeddings we need the RDF molecules i.e. the entity in a triple form e.g. $<Barrack\_Obama, wife, Michael\_Obama>$. Hence to get the corresponding RDF molecules, we are using Apache Jena, an API to create and read the Resource Descrption Framework (RDF) graphs. A sub-graph is created by extracting the RDF molecules containing the entities/relations in a given N-best sentences.

TransE~\cite{bordes2013translating} is a method to model the relationships between entities by interpreting them as translations operating on the low-dimensional embeddings of the entities. Modeling mainly means extracting the local and global connectivity pattern between entities. It is easy to train and has shown to outperform all other state-of-the art in linking multi-relational databases or knowledge bases. In this work, TransE embeddings are used to model the sub-graph of Dbpedia. 

\section{Background}

In this section, different methods used to rescore the N-best list and is relevant to the work in this paper is discussed. A query specific search result is used as a feature to rescore the n-best list in~\cite{peng2013search}. The search result contains domain specific knowledge which is difficult to capture using simple n-gram based language model. The authors have reported that with the search domain knowledge, a disambiguate query can be classified and re-ranked in the N-best list which has resulted in improved recognition accuracy. 

The standard selection criteria for a speech recognition hypothesis is maximising the posterior probability of a hypothesis $W$ given the observation sequence $X$. In~\cite{stolcke1997explicit} an algorithm to rescore N-best list is proposed, where it approximates posterior probabilities using the N-best list and then the expected word error rate is computed for each hypothesis with respect to posterior probability distribution. In~\cite{peng2013search}, various knowledge sources such as recogniser score, linguistic analysis, grammar construction, semantic discrimination score is used to rescore the N-best list. In semantic discrimination, it is only considered if the given words in a sentence are in triple form or not. Articulatory based feature is used as a knowledge source to rescore the N-best list in~\cite{li2005study}. The manner and place of articulation is used here. In~\cite{jeon2011n} the acoustic and lexical prosodic models are applied to each n-best hypothesis to obtain its prosody score, and combined with ASR scores to find the top hypothesis.

Let $w_{11}, w_{12}, w_{13},..w_{1m}..w_{1{M_1}}$ be the words in $1st-best$ sentence $s_1$ where N is the number of best sentences and M is the number of words in a sentence in the list. The recognition accuracy is usually measured for the 1st-best sentence in an $N-best$ list generated by the decoder. Most often the erroneous word existing in the 1st-best might be in correct form in the later $(N-1)-best$ list of sentences. Most of the times the word sequences are picked such that they are in some context. Since the recogniser generates the most probable word sequences which is much closer to the given test sentence, the words in the n-best list are also closer to each other e.g. it is possible that $w_{11}$ and $w_{21}$ are same words. Also words $w_{11}$, $w_{12}$ are related similarly $w_{21}$ and $w_{22}$. There are methods which uses grammar, lexicon, language rules, syntactic and semantic properties to predict a sentence with most likely word sequences similar to the test sentence. Here we are trying to see the semantic relatedness between two words.

\begin{figure}[h]
    \centering
    \resizebox{0.55\textwidth}{!}{\includegraphics{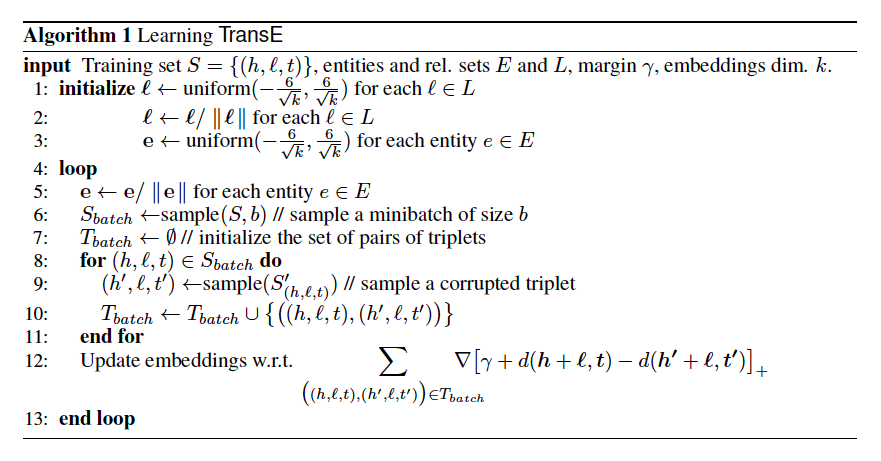}}
    \caption{TransE algorithm}
\label{fig:transe}
\end{figure}

TransE ~\cite{bordes2013translating} aims at embedding entities and relationships in a relational database (knowledge graphs) into a lower dimensional vector space. It is an energy based model. In TransE relationships are represented as translations in the embedding space: in $(h,l,t)$ the embedding of tail $t$ should be close to the embedding of head $h$ plus some vector that depends on the relationship $l$. The TransE algorithm is depicted in Figure~\ref{fig:transe}.

\section{Knowledge graph based N-best list rescoring}
\label{sec:format}

Knowledge graphs can be considered as a rich information source which can be used to interpret a sentence in logical form or in a form understandable by a computer e.g. by semantically tagging the sentence. 

\textbf{Example}: Consider a sentence, ``Chelsea Clinton is a daughter of Bill clinton and was born in Arkansas" is represented in logical/triple form as ($<Chelsea Clinton, daughter, Bill clinton>$) and  ($<Chelsea Clinton, born, Arkansas>$).  The entities in the above sentence are ``Chelsea Clinton",``Bill Clinton" and ``Arkansas", and connected with properties ``daughter" and ``born". 

It is shown in the above Example, how the entities and relations are connected in Dbpedia knowledge graph. This type of structured information storage makes it easy to retrieve the stored information and to store any new information. This type of knowledge source is useful in any language processing applications to do a logical interpretation of a sentence by understanding the entities and their relations. We are extending this knowledge graph concept to automatic speech recognition by using it to rescore the N-best list based on how the entities/relations in a sentence are semantically related. 

The steps followed for N-best list rescoring is described below.

\begin{algorithm}[H]
\floatname{algorithm}{Steps:}
\renewcommand{\thealgorithm}{}
\caption{N-best list rescoring}
\label{protocol1}
\begin{algorithmic}[1]
\STATE Obtaining the N-best list.
\STATE Annotating the mentions of Dbpedia in the N-best list
\STATE Obtaining the RDF molecules corresponding to annotated entities/relations in Step 2.
\STATE Calculating the semantic relatedness cost using TransE embeddings
\STATE Rescoring the N-best list based on the semantic relatedness cost computed in Step 4
\end{algorithmic}
\end{algorithm} 

\subsection{Obtaining the N-best list}
The N-best list of sentences are obtained from a kaldi based speech recogniser. A lattice can be defined as a labelled, weighted, directed acyclic graph (Weighted Finite State Transducers with word labels)~\cite{povey2012generating}. The N best paths are computed through the lattice using Viterbi beam search algorithm with only a single tunable parameter: the pruning beam and outputs the result as a lattice, but with a special structure. The start state in the viterbi beam search algorithm will have (up to) n arcs out of it, each one to the start of a separate path~\cite{Povey_ASRU2011}.

\subsection{Annotating the mentions of Dbpedia in the N-best list}
DBpedia Spotlight is a tool for automatically annotating mentions of DBpedia resources in text, providing a solution for linking unstructured information sources to the Linked Open Data cloud through DBpedia. DBpedia Spotlight recognizes that names of concepts or entities have been mentioned (e.g. ``Michael Jordan"), and subsequently matches these names to unique identifiers (e.g. dbpedia:Michael\_I.\_Jordan, the machine learning professor or dbpedia:Michael\_Jordan the basketball player)~\cite{daiber2013improving}. There is a web services for spotting, disambiguate and to annotate the entities/concepts. We are using only the annotation in this work. Annotation takes text as input, recognises entities/concepts to annotate and chooses an identifier in Dbpedia for each recognised entity/concept given the context.

\subsection{Obtaining the RDF molecules corresponding to annotated entities/relations in Step 2}
The entities/relations are annotated and linked to the the knowledge graph using Dbpedia spotlight. We have to see the relatedness between the entities/relations in a sentence in the graph. The enities or relations are stored only in the triple form in the graph. The entities/relation could be connected to subject or object or property of some triple combination in the graph. Let's call the entity/relations as the reference entity/relations. Using Apache Jena API, the othe triples which has the reference entity is shortlisted. The number is limited to 500. So here we get 500 RDF molecules for an entity. Like wise we obtain 500 RDF molecules for all the annotated entities/relations in a sentence.

\subsection{Calculating the semantic relatedness cost using TransE embeddings}

The connection information stored in a triple is represented in vector space using TransE~\cite{bordes2013translating}. TransE aims at embedding entities and relationships in a relational database (knowledge graphs) into a lower dimensional vector space. In TransE relationships are represented as translations in the embedding space: in $(h,l,t)$ the embedding of tail $t$ should be close to the embedding of head $h$ plus some vector that depends on the relationship $l$. If the two triples are similar then TransE encoding will have same vector values and hence the distance value will be zero. The TransE encoding takes care of the repeating subject or object or predicate in a triple. In other words the TransE encoding for the same subject in different RDF molecule will be same. The TransE embeddings are obtained for all the RDF molecules of the entities in a sentence. Here we are also using the term RDF molecule for triple. 

The semantic relatedness cost is defined as the cost function to represent the relatedness between the entities in a sentence. Although it shows the relatedness between entities in a sentence, it is derived from the RDF molecules obtained from a knowledge graph. A list of RDF molecules are obtained for an entity in a sentence. A semantic relatedness measure in the knowledge graph gives an insight into the strengths in the connection between two entities of a sentence. 

Let $M^1_t,M^2_t...M^n_t...M^N_t$ be the RDF molecule for entity  $e_t$.

%$e^1,e^2,...e^n..e^N$.

Where, $T$ is the number of entities, $N$ is the number of molecules for an entity $t$ .Here $T=500$

\begin{equation}
\delta_t = min(\mid M^n_{t+1} - M^n_t \mid),  0 \leq n \leq N-1, 
\end{equation}

\begin{equation}
\beta_m = min(\delta_t),  0 \leq t \leq T-1,
\end{equation}

Where $M$ = number of sentences and $\beta_m$ is the semantic relatedness cost.

Semantic relatedness cost is the summation of subject cost and object cost. If the distance is between subject of RDF molecule of the neighboring entities then it is called subject cost. Similarly if it is between the object of  RDF molecule then it is called the object cost. Rescoring the n-best list is done based on all these three costs (i.e. Subject\_cost, Object\_cost and total\_cost) So if smaller is the value of cost function, higher is the relatedness.

\begin{figure}[h]
    \centering
    \resizebox{0.5\textwidth}{!}{\includegraphics{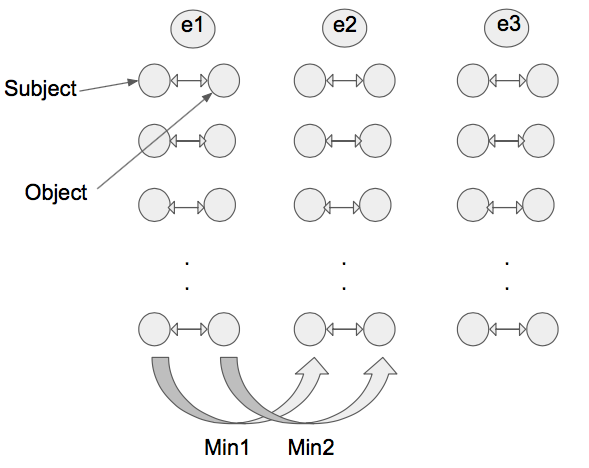}}
    \caption{Computing semantic relatedness cost using modified viterbi algorithm}
\label{fig:srkg}
\end{figure}

\subsection{Rescoring the N-best list based on the semantic relatedness cost computed in Step 4}
The semantic relatedness cost (SRC) computed for a sentence is used as a feature to rescore the N-best list. Lower the SRC, higher is the relatedness between the entities/relations in a sentence. Since the semantic relatedness cost is computed from the true information in the knowledge graph, it can be considered as a true cost.

\section{Discussion and Conclusion}
\label{expand}
A kaldi based speech recogniser is used in this work~\cite{Povey_ASRU2011}. The training database is TED-LIUM (tedlium-1) with 118hours of training date, it consists of TED talks with cleaned automatic transcripts  \footnote{http://www-lium.univ-lemans.fr/en/content/ted-lium-corpus}. Mel frequency cepstral coefficients are used as features to train a deep neural network based acoustic model and n-gram based language model. Decoding graph is created using the weighted finite state transducers~\cite{mohri2008speech}. The N-best list of sentences are obtained with $N=30$. The entity linking is made with a confidence score of 0.3 score on Dbpedia spotlight annotation tool. The Apache Jena API is used to fetch the RDF molecules from Dbpedia knowledge graph with $LIMIT=500$. TransE is trained using the RDF molecules at sentence level fetched from Dbpedia.

The above discussed method showed average results on TED-LIUM audio corpus to rescore the N best list. The method is more suitable for factoid questions with entities and relations well annotated in the N best list.

\section{Acknowledgement}
Parts of this work received funding from the European Union’s Horizon 2020 research and innovation program under the Marie Sklodowska-Curie grant agreement No. 642795 (WDAqua project).

% \begin{thebibliography}{9}
% \bibitem[1]{Davis80-COP}
%   S.\ B.\ Davis and P.\ Mermelstein,
%   ``Comparison of parametric representation for monosyllabic word recognition in continuously spoken sentences,''
%   \textit{IEEE Transactions on Acoustics, Speech and Signal Processing}, vol.~28, no.~4, pp.~357--366, 1980.
% \bibitem[2]{Rabiner89-ATO}
%   L.\ R.\ Rabiner,
%   ``A tutorial on hidden Markov models and selected applications in speech recognition,''
%   \textit{Proceedings of the IEEE}, vol.~77, no.~2, pp.~257-286, 1989.
% \bibitem[3]{Hastie09-TEO}
%   T.\ Hastie, R.\ Tibshirani, and J.\ Friedman,
%   \textit{The Elements of Statistical Learning -- Data Mining, Inference, and Prediction}.
%   New York: Springer, 2009.
% \bibitem[4]{YourName17-XXX}
%   F.\ Lastname1, F.\ Lastname2, and F.\ Lastname3,
%   ``Title of your INTERSPEECH 2017 publication,''
%   in \textit{Interspeech 2017 -- 18\textsuperscript{th} Annual Conference of the International Speech Communication Association, August 20?24, Stockholm, Sweden, Proceedings, Proceedings}, 2017, pp.~100--104.
% \end{thebibliography}

\end{document}